\documentclass[a4paper,11pt]{article}
\usepackage{jheppub} 
\usepackage{tikz} 
\usetikzlibrary{shapes,arrows,positioning,automata,backgrounds,calc,er,patterns,  shapes.geometric}

\usepackage{tikz-feynman}
\tikzfeynmanset{compat=1.1.0}
\usepackage{graphicx}

\pdfoutput=1 

\begin{document}



\title{Learning symmetries in datasets}


\author{Veronica Sanz}







\affiliation{Instituto de F\'isica Corpuscular (IFIC), Universidad de Valencia-CSIC, E-46980 Valencia, Spain}

\emailAdd{veronica.sanz@uv.es} 



\abstract{

We investigate how symmetries present in datasets affect the structure of the latent space learned by Variational Autoencoders (VAEs). By training VAEs on data originating from simple mechanical systems and particle collisions, we analyze the organization of the latent space through a relevance measure that identifies the most meaningful latent directions. We show that when symmetries or approximate symmetries are present, the VAE self-organizes its latent space, effectively compressing the data along a reduced number of latent variables. This behavior captures the intrinsic dimensionality determined by the symmetry constraints and reveals hidden relations among the features. Furthermore, we provide a theoretical analysis of a simple toy model, demonstrating how, under idealized conditions, the latent space aligns with the symmetry directions of the data manifold. We illustrate these findings with examples ranging from two-dimensional datasets with $O(2)$ symmetry to realistic datasets from electron-positron and proton-proton collisions. Our results highlight the potential of unsupervised generative models to expose underlying structures in data and offer a novel approach to symmetry discovery without explicit supervision.
 }

\maketitle

\section{Introduction}

Neural networks trained in an unsupervised fashion are not only powerful tools for compressing and generating data---they are also capable of discovering meaningful structures hidden within the data. In particular, when a dataset exhibits symmetries or approximate symmetries, a well-designed learning algorithm will often uncover these structures implicitly, using them to form more efficient internal representations. This process is at the heart of what makes unsupervised learning so appealing: without external guidance, the network develops an internal language that reflects the fundamental properties of the data.

This ability of neural networks to internalize structure has been observed across a range of domains. In our previous work \emph{Symmetry Meets AI}~\cite{Barenboim2021SymmetryAI}, we trained convolutional neural networks on a classification task involving two-dimensional potential landscapes from physics, each exhibiting a different symmetry (e.g., parity, rotational, or translational). We found that the final hidden layers of the network encoded information about the symmetry class of the input, despite the fact that symmetry was not explicitly labeled. This insight led us to define a ``symmetry score,'' which we later applied to more abstract datasets such as visual art, revealing latent symmetry-related structure even in human-made images.

In another study, \emph{Exploring How a Generative AI Interprets Music}~\cite{Barenboim2024MusicAI}, we examined the latent space of the MAGENTA generative model trained on symbolic music. We observed that the high-dimensional latent space was effectively organized around a small subset of ``relevant'' neurons, whose activations were highly correlated with musically meaningful features such as pitch, melody, and rhythm. The rest of the latent space appeared largely irrelevant to the generation process. Once again, this provided evidence that neural networks can autonomously discover the intrinsic structure of their training data---in this case, aligning with core elements of musical theory.

These findings motivate the present work, where we investigate whether a similar phenomenon occurs in the context of physical symmetries using variational autoencoders (VAEs). By training VAEs on datasets from various physical systems, we analyze whether the latent space of the model encodes the symmetries---or approximate symmetries---present in the data. 

This paper is organized as follows. In Section~\ref{sec:relatedwork}, we review previous work on symmetry discovery, equivariant models, and latent space analysis. We describe the general Variational Autoencoder architecture in Section~\ref{sec:VAE}. In Section~\ref{sec:selforganization}, we introduce the notion of self-organization in the latent space of VAEs and define the relevance measure used throughout the paper. Section~\ref{sec:toy} presents a toy model to illustrate how symmetries manifest in the latent space structure. In Section~\ref{sec:DYlep}, we apply our method to electron-positron collision data, and in Section~\ref{sec:DYhad}, to proton-proton collisions at the LHC. In Section~\ref{sec:theory}, we provide a theoretical analysis of the latent space alignment phenomenon and present a formal study of a simple toy model. Finally, we conclude in Section~\ref{sec:conclusions} with a summary of our findings and a discussion of potential extensions of this approach.

\section{Related Work}\label{sec:relatedwork}

Symmetry considerations have become central in machine learning, both for designing models and interpreting their internal representations. Many architectures bake in known invariances to improve generalization; for example, convolutional networks exploit translation symmetry by weight-sharing, and more general group-equivariant networks extend this idea to rotations and other transformations. The seminal work of Cohen and Welling~\cite{Cohen:2016} introduced group-equivariant convolutional networks, showing that encoding symmetry transformations directly into network layers can boost learning efficiency and performance. Recent geometric deep learning frameworks continue this trend, enforcing equivariance under permutations, rotations, and other group actions to build more interpretable and data-efficient models.

Beyond incorporating known symmetries into architectures, unsupervised generative models have been explored as a means to learn meaningful latent representations that capture underlying factors of variation — including symmetries — without supervision. InfoGAN~\cite{Chen2016InfoGAN} augmented GANs with an information maximization objective, showing that latent variables can spontaneously correspond to human-interpretable transformations such as digit rotation and object lighting. Around the same time, variational autoencoders (VAEs) were adapted to similar ends. The $\beta$-VAE framework~\cite{Higgins2017BetaVAE} introduced a stronger bottleneck on the latent code, encouraging more factorized representations and enabling unsupervised discovery of independent latent factors like shape, size, or orientation.

Building on these ideas, a variety of VAE variants introduced explicit regularizers for disentanglement. FactorVAE~\cite{Kim2018FactorVAE} and $\beta$-TCVAE~\cite{Chen2018TCVAE} penalize total correlation in the latent variables to encourage statistical independence, aiming to align latent dimensions with independent data factors. However, a comprehensive study by Locatello \emph{et al.}~\cite{Locatello2019Disentangle} showed that, without inductive biases or supervision, no purely unsupervised method can guarantee a fully disentangled (factorized) latent representation.

Researchers have also explored methods for discovering symmetries directly from data. Latent Space Symmetry Discovery~\cite{Yang:2024} trains generative models to map data into latent spaces where hidden symmetries become linear, uncovering nonlinear group actions underlying the dataset. In parallel, we showed in \emph{Symmetry meets AI}~\cite{Barenboim:2021} that standard feed-forward neural networks trained on symmetric inputs can implicitly encode information about the underlying symmetry in their learned representations, even without explicit labels, and applied it to art paintings.

In Particle Physics, leveraging symmetries is crucial for building interpretable and efficient models. Specialized architectures have been designed to respect known invariances of collider data. PELICAN~\cite{Bogatskiy:2022} is a graph-based network that is fully permutation-equivariant and Lorentz-invariant at each layer. Similarly, Hao \emph{et al.}~\cite{Hao:2023} introduced a Lorentz group equivariant autoencoder whose latent space respects $SO^+(3,1)$ symmetry, leading to improved performance in jet reconstruction. Self-supervised techniques like JetCLR~\cite{Dillon:2022} further leverage symmetry-based augmentations (Lorentz boosts and rotations) to learn more robust representations of particle jets.

There is also growing interest in understanding and interpreting the latent spaces learned by generative models. In general machine learning, the $\beta$-VAE framework~\cite{Higgins2017BetaVAE} and subsequent works showed that encouraging independence between latent variables often results in latent directions aligned with human-understandable features. In high-energy physics, Dillon \emph{et al.}~\cite{Dillon:2021} demonstrated that using tailored latent priors in VAEs trained on jets leads to clustered and organized latent features, improving anomaly detection.

Latent space analysis can also reveal how deep models internalize underlying symmetries. In \emph{Exploring how a generative AI interprets music}~\cite{Barenboim:2024}, we analyzed the MusicVAE~\cite{roberts2019hierarchicallatentvectormodel} and found that only a small subset of latent variables significantly impact the generated music, correlating strongly with musical concepts such as rhythm and pitch. Similar post-hoc studies, such as the analysis by Iten \emph{et al.}~\cite{Iten2020PhysicalConcepts}, suggest that deep models often gravitate toward symmetry-aligned or physically meaningful structures, even without explicit supervision.

In summary, prior work suggests a convergence of themes: building symmetries into models can make them more powerful and explainable; even without explicit guidance, neural networks often learn to encode symmetries demanded by the data; and analyzing latent spaces can reveal meaningful underlying structures.


\section{Variational Autoencoder Architecture}\label{sec:VAE}

\begin{figure}
    \centering
    \includegraphics[width=0.75\linewidth]{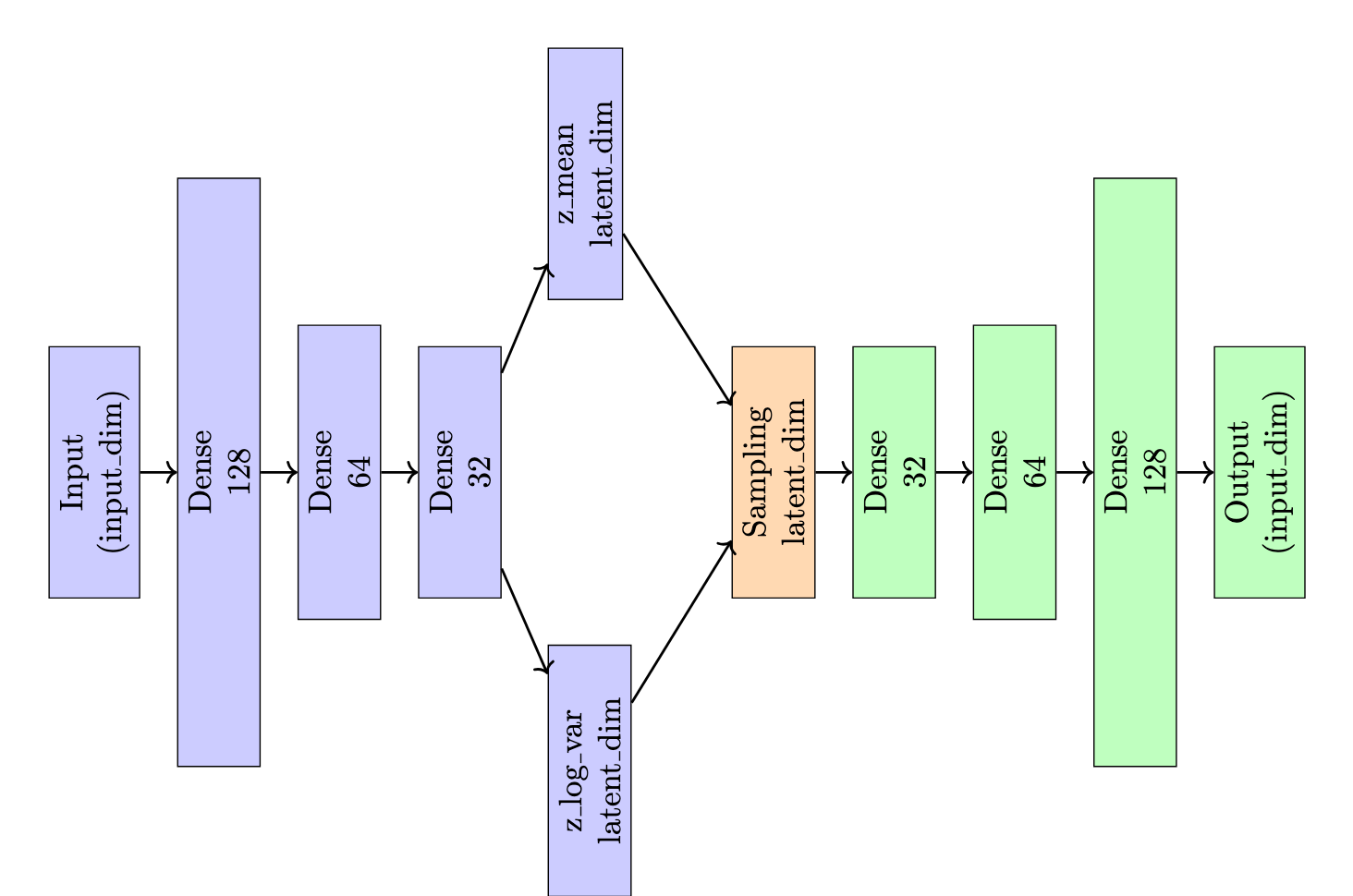}
    \caption{Variational Autoencoder architecture used in this paper. We will vary the size of  the input dimension {\tt input-dim}, and the latent dimension {\tt latent-dim} depending on the problem. }
    \label{fig:VAE}
\end{figure}

We employ a standard Variational Autoencoder (VAE) architecture consisting of an encoder and a decoder network. The structure of the network is depicted in Fig.~\ref{fig:VAE}, where the size of the input dimension is denoted by {\tt input-dim}, and the latent dimension {\tt latent-dim}. We will vary those parameters according to the dimensionality of our problem. 
The model is trained by minimizing a loss function that combines two terms: a reconstruction loss and a regularization term that enforces approximate posterior continuity. The total loss is defined as:
\begin{equation}
    \mathcal{L}_{\text{VAE}} = \mathcal{L}_{\text{rec}} + \beta \, \mathcal{L}_{\text{KL}} \ ,
\end{equation}
where $\mathcal{L}_{\text{rec}}$ is the mean squared error (MSE) between the input and reconstructed output, and $\mathcal{L}_{\text{KL}}$ is the Kullback-Leibler (KL) divergence between the encoded latent distribution and a standard normal prior. Explicitly, the loss reads:
\begin{equation}
\begin{aligned}
    \mathcal{L}_{\text{rec}} &= \frac{1}{N} \sum_{i=1}^{N} \left\| \mathbf{x}^{(i)} - \hat{\mathbf{x}}^{(i)} \right\|^2 \ , \\\\
    \mathcal{L}_{\text{KL}} &= -\frac{1}{2N} \sum_{i=1}^{N} \sum_{j=1}^{d_z} \left(1 + \log \sigma_j^{2(i)} - \mu_j^{2(i)} - \sigma_j^{2(i)} \right) \ ,
\end{aligned}
\end{equation}
where $\mathbf{x}^{(i)}$ and $\hat{\mathbf{x}}^{(i)}$ are the input and reconstructed data points, and $\mu_j^{(i)}$ and $\sigma_j^{(i)}$ are the mean and standard deviation of the latent variable $z_j$ for event $i$. The hyperparameter $\beta$ controls the weight of the KL term and is set to $\beta = 0.1$ in our training.

The model is optimized using the Adam algorithm with a learning rate of $10^{-3}$.
For the datasets discussed in this work, it is sufficient to run the VAE for 15-30 epochs.


\section{Self-Organization in the Latent Space}\label{sec:selforganization}

A central observation in our analysis is that, when trained on data that exhibit underlying symmetries or constraints, the latent space of a Variational Autoencoder (VAE) tends to \emph{self-organize}, aligning along the truly independent degrees of freedom of the dataset. This behavior can be interpreted as the network's way of adapting to the intrinsic structure of the data: symmetries and constraints reduce the effective dimensionality of the system, and the VAE spontaneously reflects this reduction in its latent variables.

To illustrate this phenomenon, we begin with a simple toy example~\ref{sec:toy}. We compare two situations by constructing two-dimensional datasets where the variables $x_1$ and $x_2$ are either statistically independent or related by a symmetry constraint. When the dataset consists of points sampled independently in $x_1$ and $x_2$, we observe that the relevance distribution of the latent neurons ----quantified by the variance of each latent coordinate in the dataset ---- is flat, with two latent variables clearly dominating and the remaining dimensions showing negligible relevance. This is expected as the data have two degrees of freedom.

However, when the dataset is constrained to lie on a circle ---- a simple symmetry relation between $x_1$ and $x_2$ ----the relevance distribution exhibits a qualitatively different behavior. In this case, we observe a single dominant latent dimension, followed by a steep drop in the relevance of the remaining neurons. This indicates that the VAE has effectively identified that the data live on a one-dimensional manifold embedded in two dimensions and has reorganized its latent space accordingly. The relevance distribution thus provides a robust and interpretable measure of the true dimensionality of the system, capturing the presence of symmetries or constraints in the data.

We extend this analysis to a more realistic scenario from particle physics: electron-positron collisions producing muon-antimuon pairs, known as Drell-Yan scattering at fixed center-of-mass energy. In this case, the dataset consists of six variables corresponding to the three-momenta of the final-state muons. When training a VAE on this dataset, we observe that the relevance distribution is sharply peaked around three latent dimensions, with the remaining dimensions contributing negligibly. This is consistent with the known kinematic constraints of the process: in the laboratory frame, momentum conservation imposes three conditions on the momenta of the final-state particles, effectively reducing the dimensionality of the system from six to three. Further analysis reveals that activations of the three relevant latent neurons correlate strongly with specific combinations of muon and antimuon momenta, in particular with differences $\bf p_{\mu} - \bf p_{\bar{\mu}}$. This provides clear evidence that the VAE has internalized the physical constraint of momentum conservation and reorganized its latent space to reflect the true degrees of freedom of the process. We then extend this discussion to a more complex situation, dimuon production at the Large Hadron Collider (LHC) where the energy of each collision is not fixed and both virtual photons and massive $Z$ bosons contribute to the production.


\subsection{Toy Example: A Two-Dimensional Dataset with and without Symmetry}\label{sec:toy}

To illustrate how the latent space of a VAE self-organizes in response to the intrinsic dimensionality and symmetries of the data, we start with a simple two-dimensional example. We construct two datasets:

\begin{itemize}
    \item A truly two-dimensional dataset with independent variables:
    \begin{equation}
        \mathcal{D}^{\text{2D}} = \left(x_1^i, x_2^i\right)
    \end{equation}
    where $x_1$ and $x_2$ are sampled independently from a uniform distribution in the interval $[-R, R]$.
    
    \item A dataset where the two variables are constrained by an $O(2)$ symmetry:
    \begin{equation}
        \mathcal{D}^{\text{1D}} = \left(x_1^i, x_2^i\right) \quad \text{with} \quad x_1^2 + x_2^2 = R^2 \ ,
    \end{equation}
    which corresponds to points distributed uniformly on a circle of fixed radius $R$.
\end{itemize}

For both datasets, we generate $10^4$ events with $R = 10$.
We train a Variational Autoencoder (VAE) with a four-dimensional latent space on each of these datasets, based on the architecture in Sec.~\ref{sec:VAE}.

After training, we analyze the structure of the latent space by computing the \emph{relevance} of each latent dimension.

We define the relevance $\rho_j$ of the $j$-th latent variable as:
\begin{equation}\label{eq:relevance}
    \textrm{Relevance}_j = \frac{\text{std}(   \langle z_j \rangle  )}{\text{mean} (\sigma_j)} \ ,
\end{equation}
where we compute the ration between the standard deviation of the $j$-th latent coordinate over the dataset and the mean of $\sigma_j$, the variance within each event. 

This definition was proposed by our study in Ref.\cite{Barenboim2024MusicAI}, where we found that this ratio was a good criteria for classifying the relevant directions in the latent space of the MusicVAE\footnote{Note, however, alternative relevance measures have been proposed in the literature. In particular, total-correlation-regularized VAEs (TC-VAEs) such as FactorVAE and $\beta$-TCVAE explicitly encourage latent independence by penalizing the total correlation (TC) among latent dimensions \citep{Kim2018,Chen2018}. This TC-based notion of relevance is intrinsically tied to disentanglement objectives, since reducing total correlation encourages each latent dimension to capture an independent factor of variation.}.
\begin{figure}[t!]
    \centering
    \includegraphics[width=0.7\linewidth]{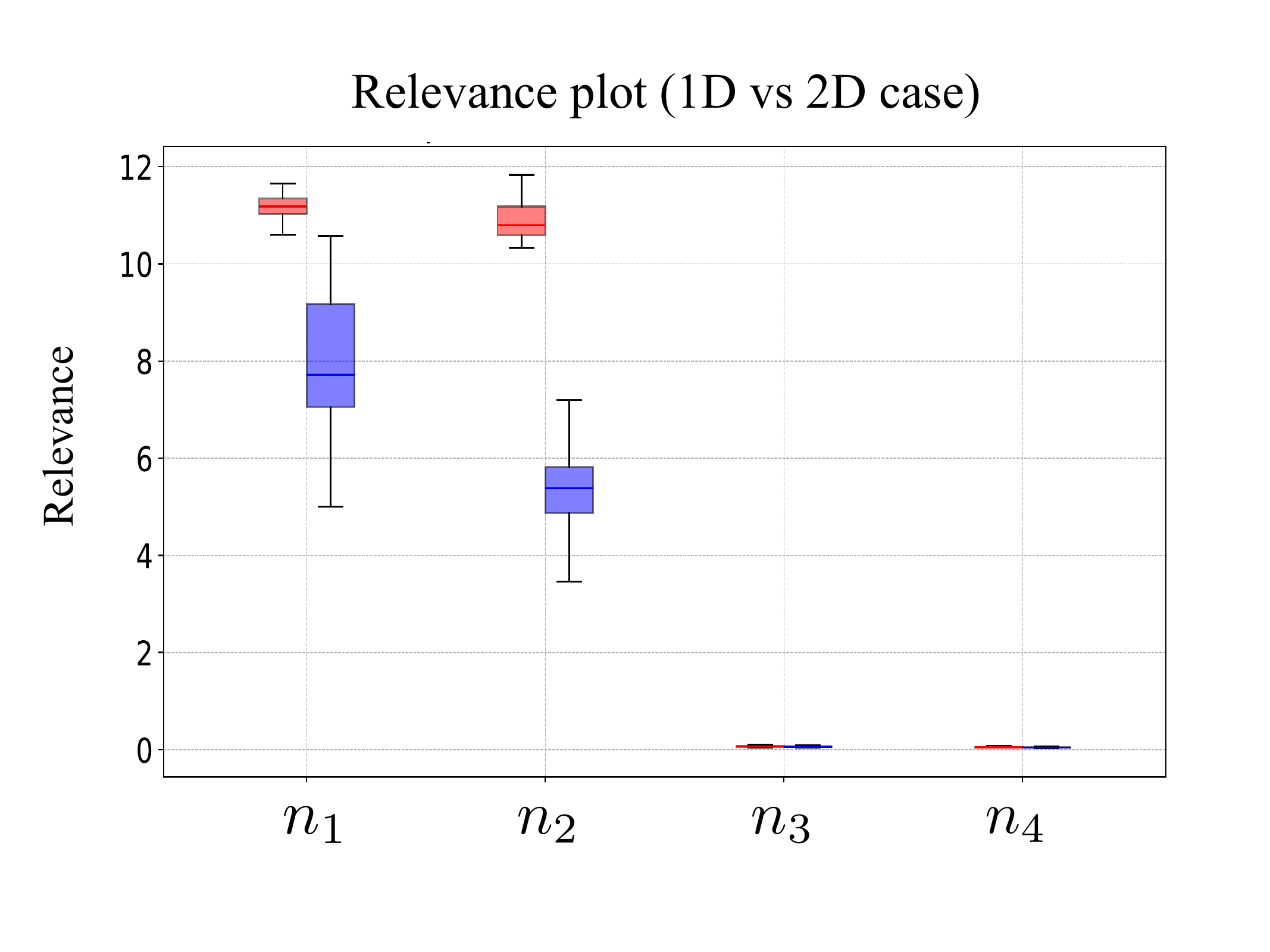}
    \caption{Relevance distribution of the latent variables for the two datasets. In orange, the truly two-dimensional dataset $\mathcal{D}^{\text{2D}}$ and in blue the dataset constrained to a circle $\mathcal{D}^{\text{1D}}$. The latent variables are ordered by decreasing relevance.}
    \label{fig:relevance}
\end{figure}

Figure~\ref{fig:relevance} shows the relevance distribution for both datasets after running 20 times a VAE in each dataset. In the case of the truly two-dimensional dataset $\mathcal{D}^{\text{2D}}$, we observe that two latent variables exhibit high and comparable relevance, while the other two show negligible relevance. This indicates that the VAE has identified the intrinsic dimensionality of the data, effectively discarding unnecessary latent directions.

In contrast, for the dataset constrained by the $O(2)$ symmetry, $\mathcal{D}^{\text{1D}}$, the relevance plot displays a clear hierarchy: a single latent dimension has dominant relevance, followed by a steep decline. This reflects the fact that the data lives on a one-dimensional manifold embedded in two dimensions, and the VAE reorganizes its latent space accordingly.

\begin{figure}[h!]
    \centering
    \includegraphics[width=0.48\textwidth]{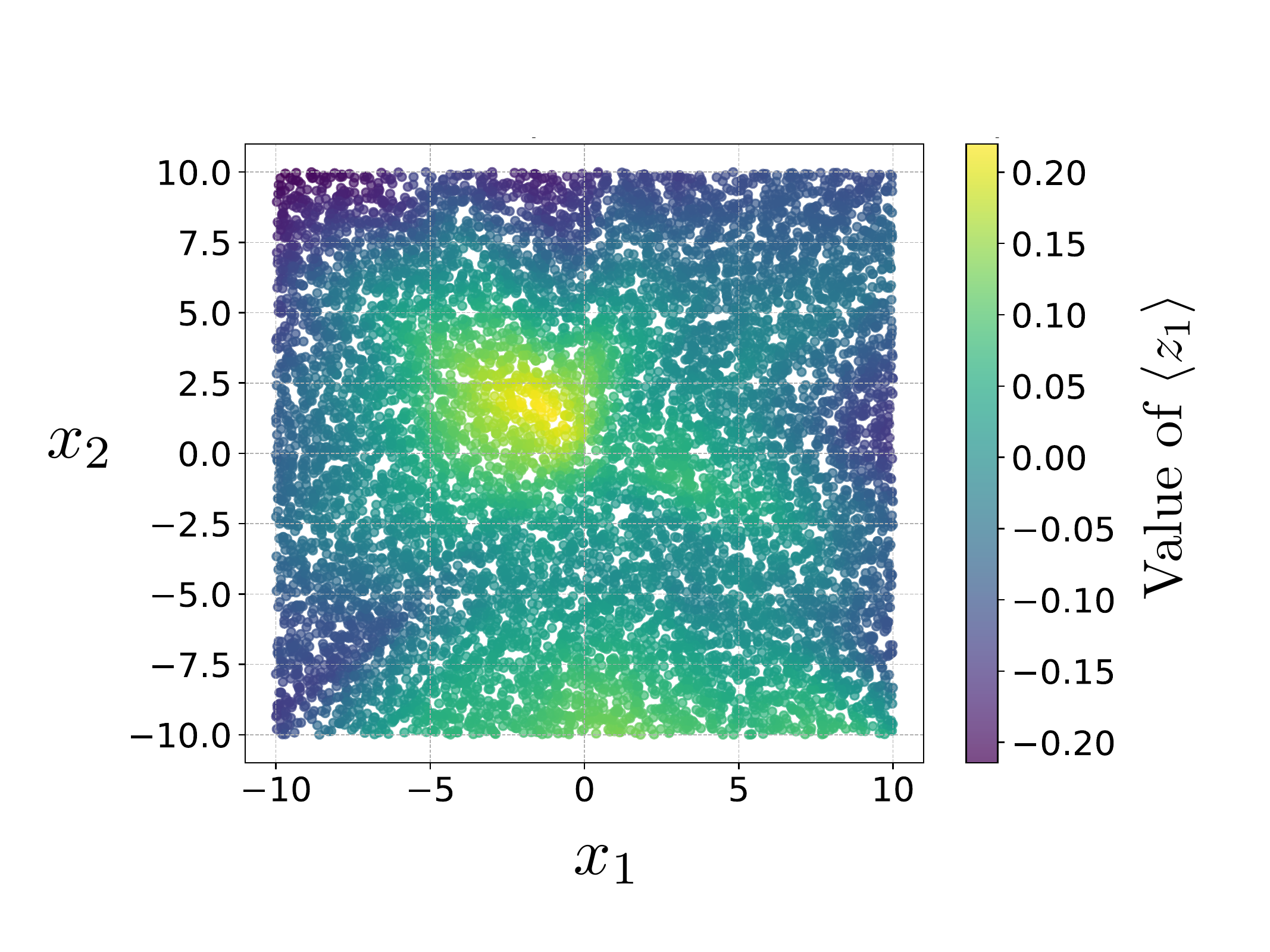}
    \includegraphics[width=0.48\textwidth]{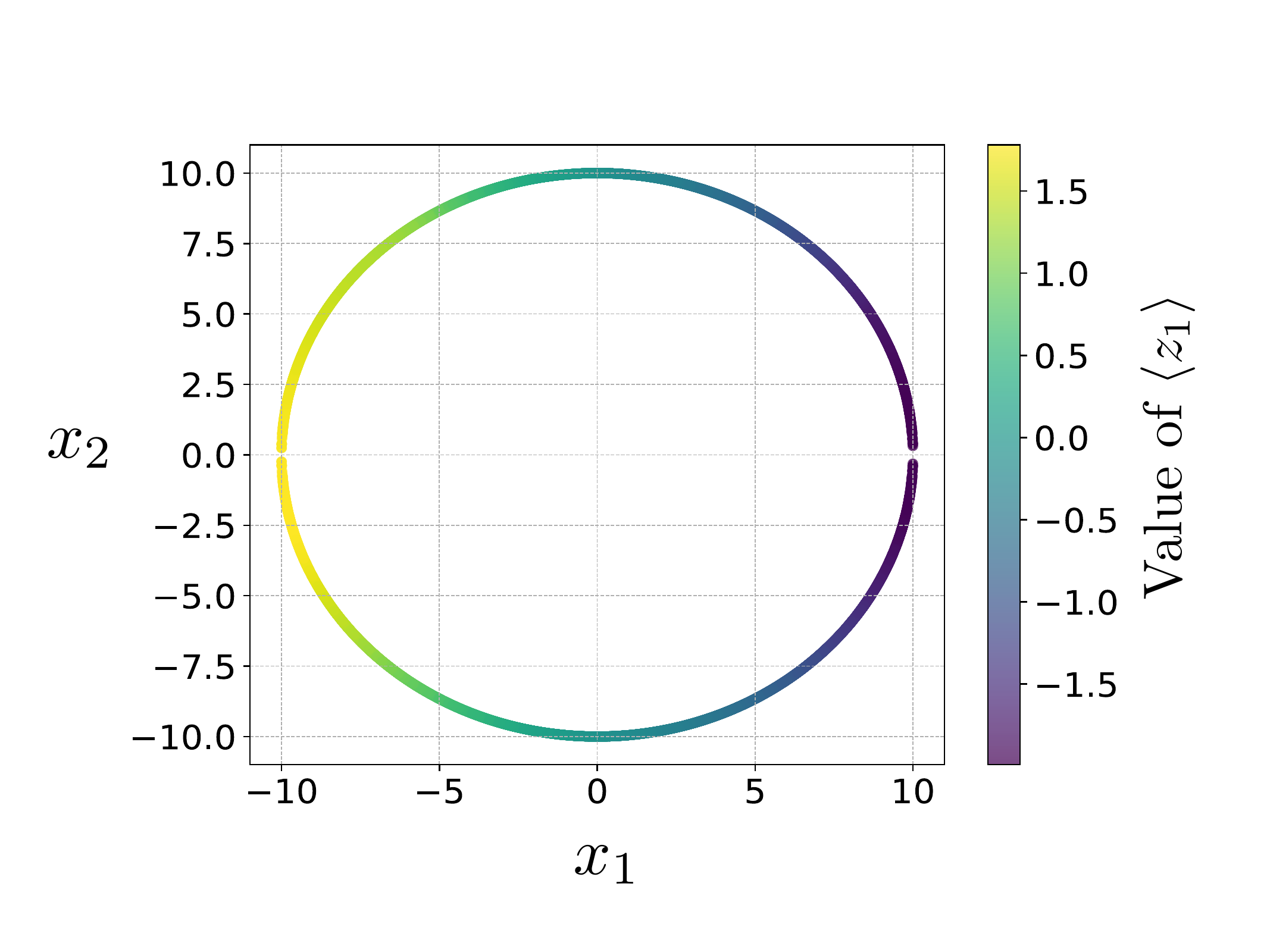}
    \caption{Distribution of the mean latent activation $\langle z_1 \rangle$ (most relevant latent direction) in the $(x_1, x_2)$ space. Left: truly two-dimensional dataset $\mathcal{D}^{\text{2D}}$. Right: $O(2)$ symmetric dataset $\mathcal{D}^{\text{1D}}$. In the symmetric case, the latent variable is ordered along the circle.}
    \label{fig:zmean_distribution}
\end{figure}

In addition to the relevance distribution, we examine how the most relevant latent variable organizes over the input space. In Figure~\ref{fig:zmean_distribution}, we show the distribution of the mean latent activation $\langle z_1 \rangle$ (corresponding to the most relevant latent direction) projected onto the $(x_1, x_2)$ space, for both datasets. In the truly two-dimensional dataset $\mathcal{D}^{\text{2D}}$, we observe no particular structure: the values of $\langle z_1 \rangle$ are scattered uniformly without any clear pattern. This reflects the lack of constraints relating $x_1$ and $x_2$ in this case.

In contrast, for the $O(2)$ symmetric dataset $\mathcal{D}^{\text{1D}}$, the projection reveals a different structure. As expected, the data points lie on a circle in the $(x_1, x_2)$ plane. Moreover, the values of $\langle z_1 \rangle$ are continuously ordered along the circle, indicating that the VAE has learned to parametrize the underlying one-dimensional manifold. This demonstrates how the VAE reorganizes its latent space to capture not only the dimensionality reduction induced by the symmetry but also the geometric structure of the data.

\begin{figure}[h!]
    \centering
    \includegraphics[width=0.48\textwidth]{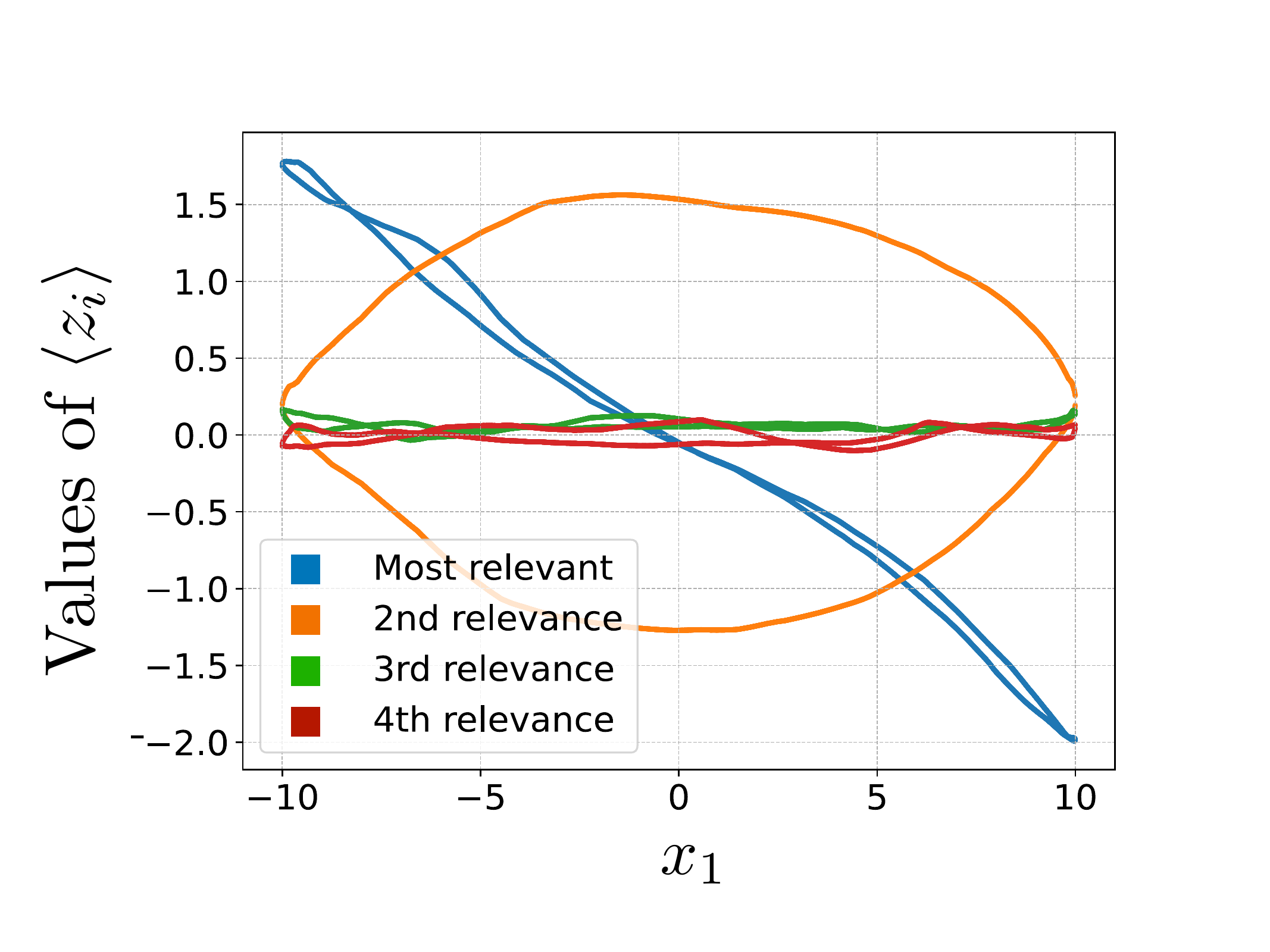}
    \includegraphics[width=0.48\textwidth]{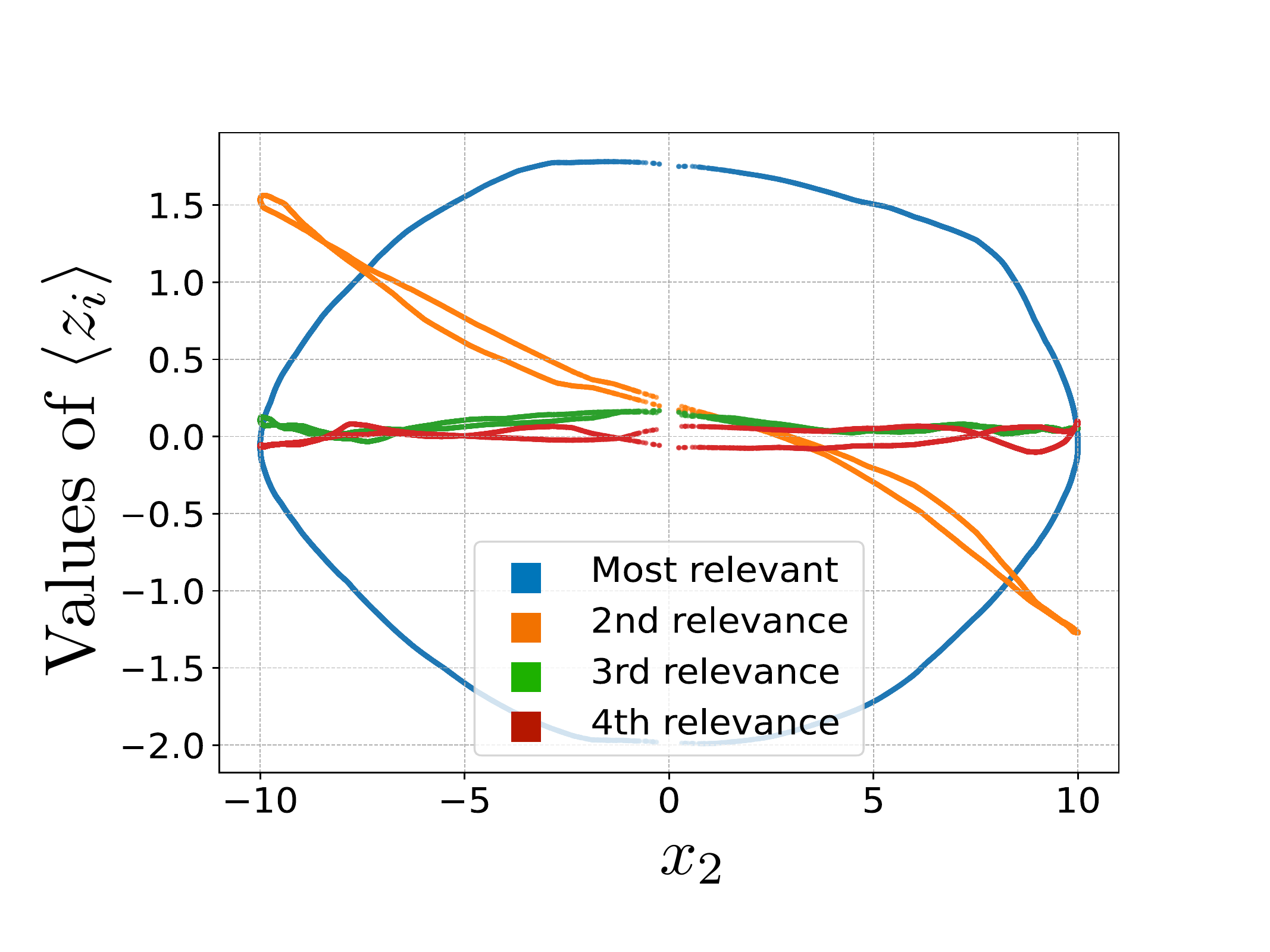}
    \caption{Mean latent activations for all the latent dimensions in the $O(2)$ symmetric datase, $\langle z_i \rangle$, $i=1\ldots4$ as a function of $x_1$ (left) and $x_2$ (right).}
    \label{fig:latent_structure}
\end{figure}

To further investigate the structure of the latent space, we analyze the mean latent activations  as a function of the input features $x_1$ and $x_2$. Figure~\ref{fig:latent_structure} shows scatter plots of the two most relevant latent variables for the $O(2)$ symmetric dataset. We observe that the first latent variable is linearly correlated with $x_1$ and exhibits an elliptical structure when plotted against $x_2$. Conversely, the second latent variable shows a similar behavior but with $x_1$ and $x_2$ interchanged. This reveals that the VAE has organized its latent space to align with the natural variables of the dataset, effectively encoding the symmetry relation between $x_1$ and $x_2$.

This toy example clearly demonstrates that the latent space of a VAE self-organizes to reflect the intrinsic structure and symmetries of the data. The relevance distribution offers a quantitative measure of the effective dimensionality of the system, while the structure of the latent variables reveals how the model internalizes the relationships among the data features.

\subsection{Application to Lepton Collisions}\label{sec:DYlep}

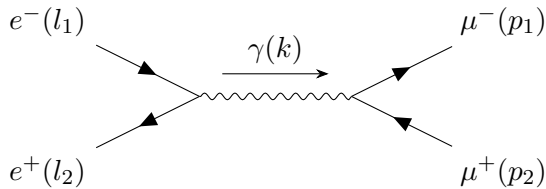
\begin{figure}[h!]
\centering
\begin{tikzpicture}
  \begin{feynman}
    \vertex (e-) at (-2,1) {\(e^-(l_1)\)};
    \vertex (e+) at (-2,-1) {\(e^+(l_2)\)};
    \vertex (v1) at (0,0);
    \vertex (v2) at (2,0);
    \vertex (mu-) at (4,1) {\(\mu^-(p_1)\)};
    \vertex (mu+) at (4,-1) {\(\mu^+(p_2)\)};
    
    \diagram* {
      (e-) -- [fermion] (v1) -- [fermion] (e+),
      (v1) -- [photon, momentum=\(\gamma (k)\)] (v2),
      (v2) -- [fermion] (mu-),
      (mu+) -- [fermion] (v2),
    };
  \end{feynman}
\end{tikzpicture}
\caption{Feynman diagram for the process \( e^+ e^- \to \mu^+ \mu^- \) in QED.}
\label{fig:eemumu}
\end{figure}

We now turn to a more realistic application of our method, using data from a well-understood physical process: electron-positron collisions producing a muon-antimuon pair in Quantum Electrodynamics (QED). We generate simulated events describing the kinematics of the process:
\begin{equation}
    e^+ e^- \rightarrow \mu^+ \mu^- \ .
\end{equation}
whose Feynman diagram is shown in Fig.~\ref{fig:eemumu}.
\begin{figure}[t!]
    \centering
    \includegraphics[width=0.6\textwidth]{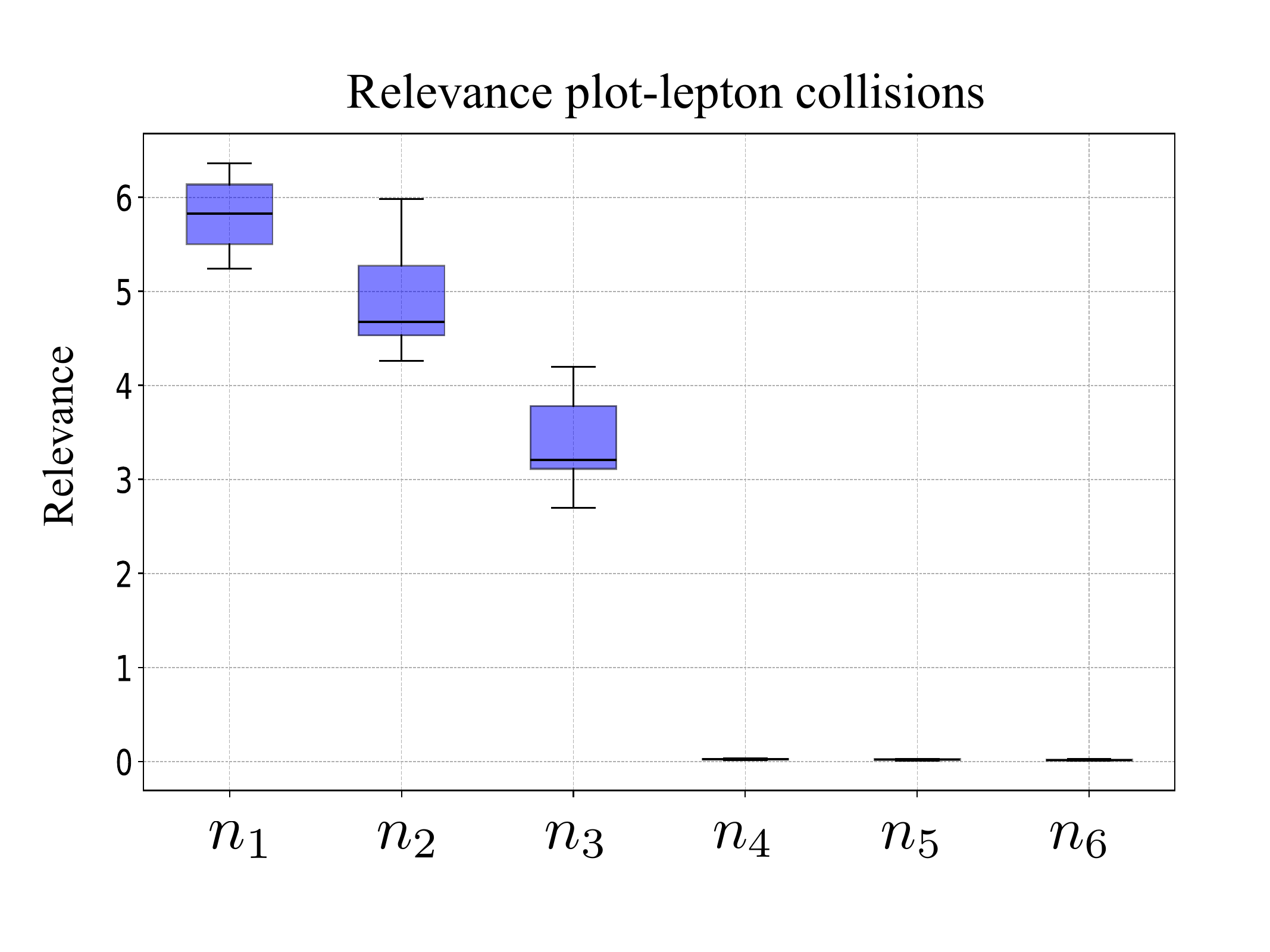}
    \caption{Relevance distribution of the latent variables for the Drell-Yan dataset. Only three latent dimensions exhibit significant relevance, reflecting the physical constraint of momentum conservation.}
    \label{fig:relevance_dy}
\end{figure}
Each event is generated at fixed center-of-mass energy, $\sqrt{s} = 80$~GeV using {\tt MadGraph5\_aMC@NLO}~\cite{Alwall:2014hca,Frederix:2018nkq}.

The observable quantities in this process are the four-momenta of the muon and antimuon. As input features for the VAE, we consider the three-momenta of each of the two muons in the laboratory frame:
\begin{equation}
    \mathcal{D}^{\text{DY}} = (\mathbf{p}_\mu, \mathbf{p}_{\bar{\mu}}) \ ,
\end{equation}
where $\mathbf{p}_\mu = (p_{x}^{\mu}, p_{y}^{\mu}, p_{z}^{\mu})$ and $\mathbf{p}_{\bar{\mu}} = (p_{x}^{\bar{\mu}}, p_{y}^{\bar{\mu}}, p_{z}^{\bar{\mu}})$.

We train a Variational Autoencoder with a six-dimensional latent space on this dataset. As in the toy example, we analyze the structure of the latent space by computing the relevance of each latent variable, using the same definition as in Eq.~\ref{eq:relevance}.

The resulting relevance distribution is shown in Figure~\ref{fig:relevance_dy}. We observe that only three latent directions exhibit significant relevance, while the other three are negligible. This is a clear indication that the VAE has identified the effective dimensionality of the system as three, despite being trained on six-dimensional input data.

This reduction is a direct consequence of momentum conservation in the laboratory frame. Since the initial state is at rest and has no net momentum, the total three-momentum of the final-state muon-antimuon pair must vanish:
\begin{equation}
\begin{aligned}
    p_{x}^{\mu} + p_{x}^{\bar{\mu}} &= 0 \ , \\
    p_{y}^{\mu} + p_{y}^{\bar{\mu}} &= 0 \ , \\
    p_{z}^{\mu} + p_{z}^{\bar{\mu}} &= 0 \ .
\end{aligned}
\label{eq:momentum_conservation}
\end{equation}
These three conditions reduce the number of independent degrees of freedom from six to three.

\begin{figure}[ht!]
    \centering
    \includegraphics[angle=0,width=0.8\textwidth]{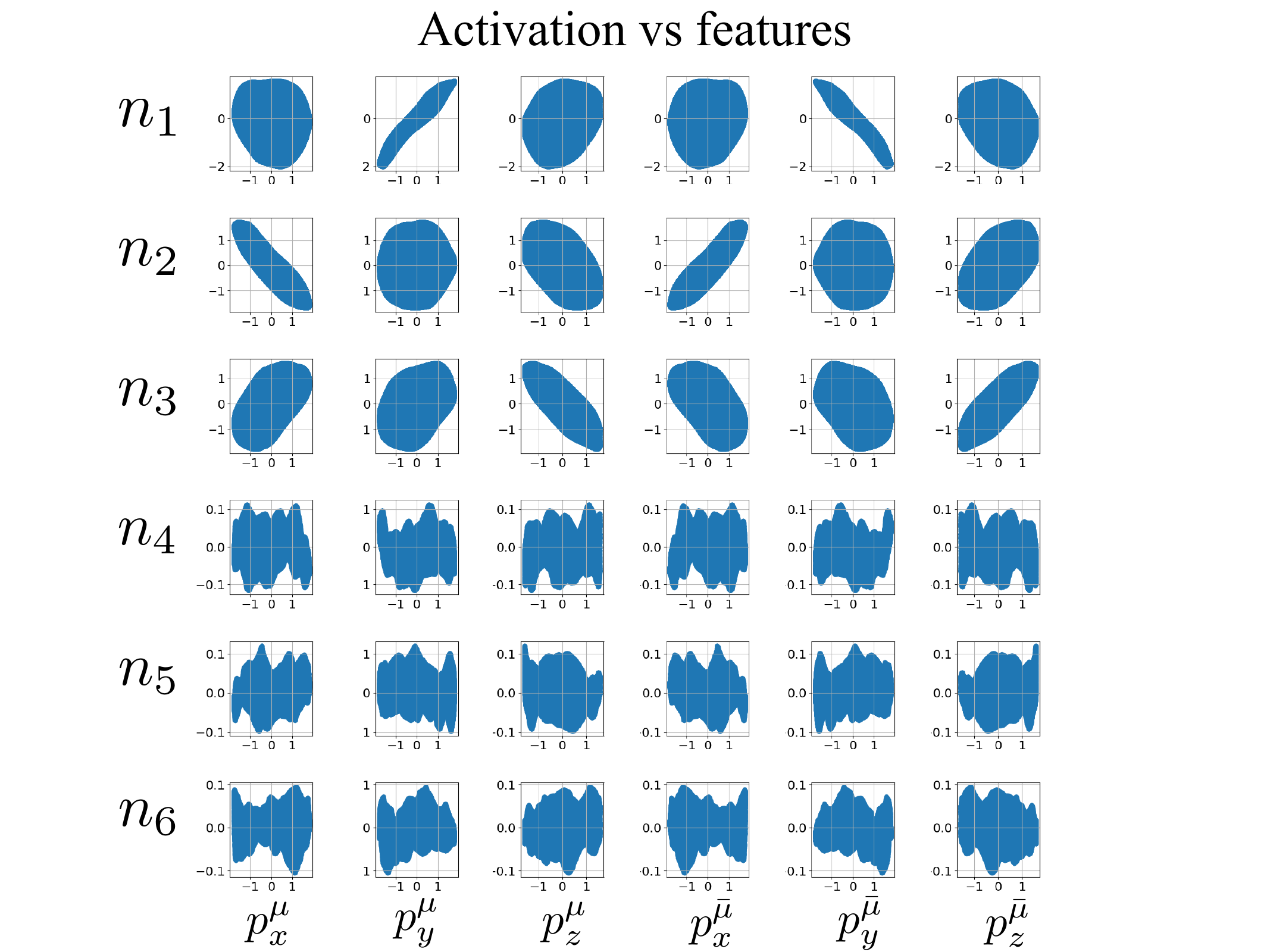}
    \caption{Array of scatter plots showing the mean latent activations  as a function of the six input features, for the Drell-Yan dataset. The first three latent variables exhibit strong correlations with $p_{y}^{\mu} - p_{y}^{\bar{\mu}}$, $p_{x}^{\mu} - p_{x}^{\bar{\mu}}$, and $p_{z}^{\mu} - p_{z}^{\bar{\mu}}$, respectively. The remaining latent variables show no significant correlation with the input features.}
    \label{fig:zmean_dy}
\end{figure}

To further investigate how this structure is encoded in the latent space, we analyze the mean activations  of each latent variable as a function of the six input features. Figure~\ref{fig:zmean_dy} shows a $6 \times 6$ array of scatter plots, displaying the correlation between $\langle z_j \rangle$ and each of the input components.
A clear pattern emerges: the three relevant latent directions exhibit strong correlations with specific linear combinations of the input features. In particular, we observe that:
\begin{equation}
\begin{aligned}
    \langle z_2 \rangle &\sim p_{x}^{\mu} - p_{x}^{\bar{\mu}} \ , \\
    \langle z_1 \rangle &\sim p_{y}^{\mu} - p_{y}^{\bar{\mu}} \ , \\
    \langle z_3 \rangle &\sim p_{z}^{\mu} - p_{z}^{\bar{\mu}} \ .
\end{aligned}
\end{equation}
Conversely, the mean activations of the remaining latent variables $\langle z_4 \rangle$, $\langle z_5 \rangle$, and $\langle z_6 \rangle$ show no discernible correlation with any of the input features, consistent with their low relevance.

These results demonstrate that the VAE has autonomously identified and encoded the fundamental kinematic constraint of three-dimensional momentum conservation in its latent space. The relevance distribution and the structure of the latent activations provide clear evidence that the model has discovered and internalized the symmetry of the system without explicit supervision.

\subsection{Application to Drell-Yan Production in Hadron Collisions}
\label{sec:DYhad}

We further investigate the self-organization of the latent space in the case of Drell-Yan production of muon pairs in proton-proton collisions at a center-of-mass energy of 13~TeV. The process is described at partonic level by:
\begin{equation}
    q + \bar{q} \rightarrow \gamma^*, Z \rightarrow \mu^+ + \mu^- \ .
\end{equation}

The observable quantities in this case are the four-momenta of the muon and antimuon:
\begin{equation}
    \mathcal{D}^{\text{DY}} = \left(p^{\mu}_\mu, p^{\mu}_{\bar{\mu}} \right) \ ,
\end{equation}
which constitute an eight-dimensional dataset.

A key difference with respect to electron-positron collisions is that in hadronic collisions, the initial state is not fully known. The incoming partons (quark and antiquark) carry unknown momentum fractions  of the incoming protons, and thus the longitudinal momentum and energy of the partonic system fluctuate event by event. However, the transverse momentum of the initial state is negligible, leading to approximate transverse momentum conservation:
\begin{equation}
\begin{aligned}
    p_{x}^{\mu} + p_{x}^{\bar{\mu}} &\simeq 0 \ , \\\\
    p_{y}^{\mu} + p_{y}^{\bar{\mu}} &\simeq 0 \ .
\end{aligned}
\label{eq:transverse_momentum}
\end{equation}

In addition, the final-state muons are on-shell particles, satisfying:
\begin{equation}
    p_{\mu}^2 = m_{\mu}^2 \ , \quad p_{\bar{\mu}}^2 = m_{\mu}^2 \ .
\end{equation}
At high energies, the muon mass is negligible compared to the energy scale of the collision, so we can approximate:
\begin{equation}
    p_{\mu}^2 \simeq 0 \ , \quad p_{\bar{\mu}}^2 \simeq 0 \ .
\end{equation}





At the LHC and with selection cuts for the muons, the  Drell-Yan process is dominated by an on-shell $Z$ boson. In this case, an additional constraint is imposed on the system: the invariant mass of the muon-antimuon pair is fixed to the $Z$ boson mass,
\begin{equation}
    (p_{\mu} + p_{\bar{\mu}})^2 = m_Z^2 \ .
\end{equation}

Including this condition, the total number of kinematic constraints increases to five.

Therefore, the effective dimensionality of the system is  reduced to:
\begin{equation}
    8 \ (\text{observables}) - 5 \ (\text{constraints}) = 3 \ (\text{degrees of freedom}) \ .
\end{equation}

This scenario is directly comparable to the electron-positron collision example discussed previously, where the kinematic constraints reduced the dimensionality of the dataset to three.

\begin{figure}[h!]
    \centering
    \includegraphics[width=0.75\linewidth]{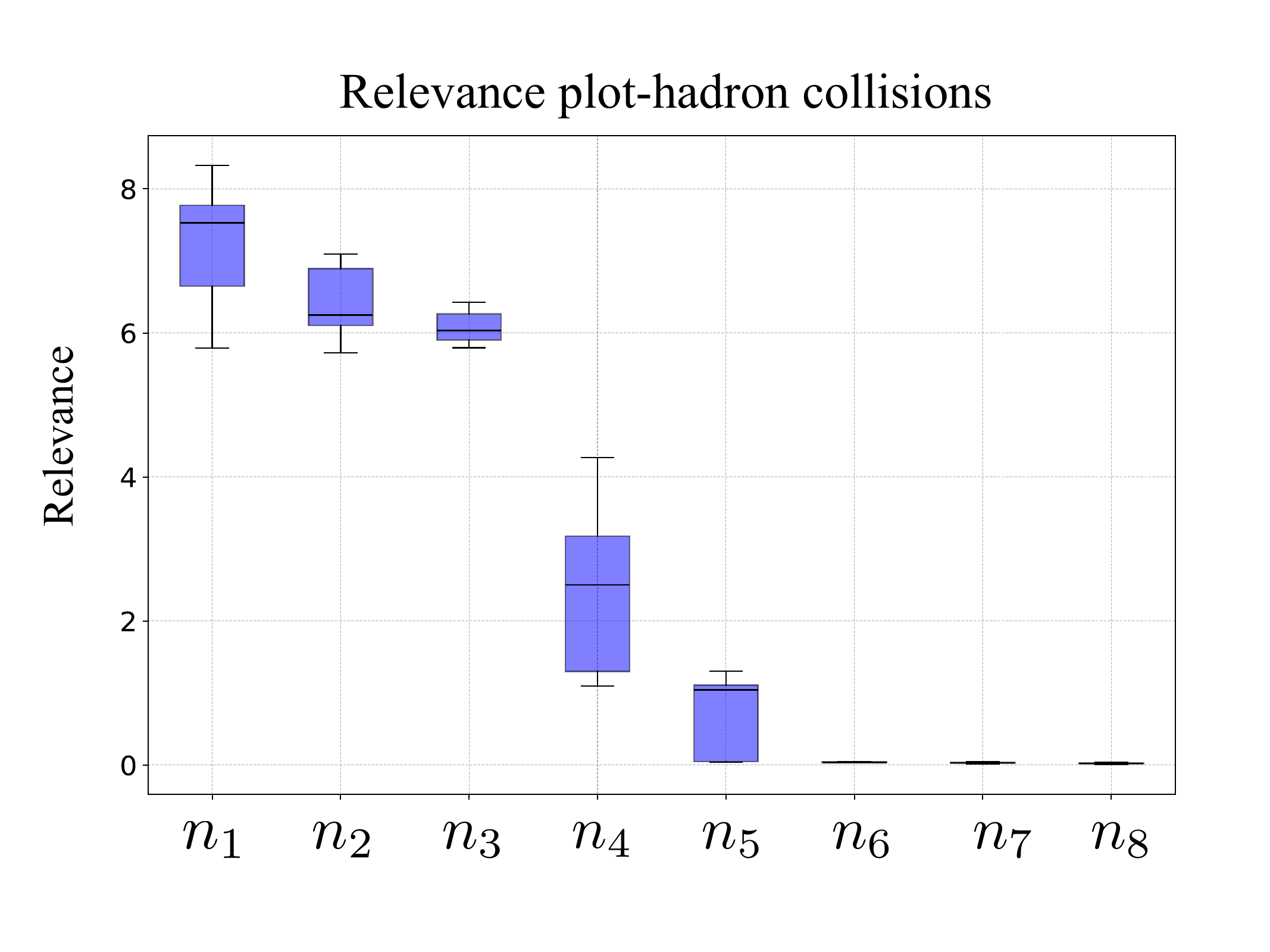}
    \caption{Relevance distribution of the latent variables for the Drell-Yan dataset in proton-proton collisions at $\sqrt{s} = 13$~TeV. The latent variables are ordered by decreasing relevance. The plot shows a clear hierarchy, with three latent dimensions exhibiting significantly higher relevance, in agreement with the expected effective dimensionality imposed by kinematic constraints.}
    \label{fig:relevancepp}
\end{figure}

We produce events $p \, p \to \mu^+ \mu^-$ using {\tt MadGraph5\_aMC@NLO}~\cite{Alwall:2014hca,Frederix:2018nkq} at LHC energies of 13 TeV and train a VAE with 8 input features (four-momenta of the muon and antimuon). The structure of the latent space learned by the VAE reflects the kinematic constraints discussed before.  In Figure~\ref{fig:relevancepp}, we show the relevance distribution of the latent variables, ordered by decreasing relevance and averaged over 15 runs. The relevance is computed according to the same criterion defined in Equation~\ref{eq:relevance}, providing a quantitative measure of the information content encoded in each latent dimension. The plot exhibits a clear hierarchy: the first three latent variables display significantly higher relevance compared to the remaining ones, which rapidly fall to negligible values. This behavior indicates that the VAE has effectively identified the intrinsic dimensionality of the system as three, consistent with the known physical constraints in the case of on-shell $Z$ production. 

\section{On the Alignment of Latent Spaces with Symmetry Directions}\label{sec:theory}

The results presented in this paper suggest that Variational Autoencoders (VAEs) trained on datasets exhibiting symmetries tend to organize their latent spaces along directions aligned with the symmetry transformations. However, it is important to emphasize that no general mathematical proof guarantees this behavior for arbitrary datasets and standard VAE architectures.

In a standard VAE, the objective is to maximize the evidence lower bound (ELBO),
\begin{equation}
\mathcal{L}_{\text{VAE}} = \mathbb{E}_{q(\mathbf{z}|\mathbf{x})} [\log p(\mathbf{x}|\mathbf{z})] - \beta \, D_{\mathrm{KL}}(q(\mathbf{z}|\mathbf{x}) \parallel p(\mathbf{z})) \ ,
\end{equation}
where $q(\mathbf{z}|\mathbf{x})$ is the encoder, $p(\mathbf{x}|\mathbf{z})$ the decoder, and $p(\mathbf{z})$ the prior distribution. This loss function trades off reconstruction fidelity and prior matching, but it does not impose any explicit constraint aligning latent variables with symmetries in the dataset. Consequently, without additional assumptions or architectural modifications, there is no general theoretical guarantee that the latent space will organize along symmetry directions.

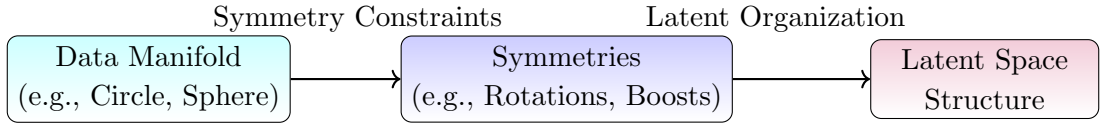
\begin{figure}[ht]
    \centering
    \begin{tikzpicture}[node distance=2.5cm, every node/.style={align=center}]
    
    \node (data) [rectangle, draw, rounded corners, minimum width=3cm, minimum height=1cm,
        top color=cyan!20, bottom color=white, shading=axis] {Data Manifold \\ (e.g., Circle, Sphere)};
        
    \node (symmetry) [rectangle, draw, rounded corners, right of=data, minimum width=3cm, minimum height=1cm, xshift=3cm,
        top color=blue!20, bottom color=white, shading=axis] {Symmetries \\ (e.g., Rotations, Boosts)};
        
    \node (latent) [rectangle, draw, rounded corners, right of=symmetry, minimum width=3cm, minimum height=1cm, xshift=3cm,
        top color=purple!20, bottom color=white, shading=axis] {Latent Space \\ Structure};

    \draw[->, thick] (data) -- (symmetry);
    \draw[->, thick] (symmetry) -- (latent);

    \node at ($(data)!0.5!(symmetry)$) [above=0.5cm] {Symmetry Constraints};
    \node at ($(symmetry)!0.5!(latent)$) [above=0.5cm] {Latent Organization};

    \end{tikzpicture}
    \caption{Schematic illustration of how symmetries present in the data manifold influence the organization of the latent space learned by a Variational Autoencoder (VAE).}
    \label{fig:schematic_symmetries}
\end{figure}

Nevertheless, several lines of research provide partial theoretical support for this phenomenon under more restrictive conditions. In the case of linear models, such as linear autoencoders, it is well known that the principal components of the data — as recovered through Principal Component Analysis (PCA)~\cite{Jolliffe2002PCA,Baldi1989LinearAE} — align with directions of maximal variance, which often correspond to symmetry axes when the data distribution is symmetric. This behavior can be seen as a linear analog of the latent space organization observed in VAEs.

Furthermore, in the context of disentangled representation learning, models such as $\beta$-VAE~\cite{Higgins2017BetaVAE}, FactorVAE~\cite{Kim2018FactorVAE}, and $\beta$-TCVAE~\cite{Chen2018TCVAE} introduce regularization terms that encourage statistical independence between latent variables. These methods promote a factorized latent representation, where ideally each latent coordinate corresponds to an independent factor of variation. Although these models improve disentanglement, a rigorous study by Locatello \emph{et al.}~\cite{Locatello2019Disentangle} proved that purely unsupervised disentanglement is fundamentally impossible without inductive biases or supervision. Thus, while encouraging independent latent factors may help align with symmetry directions, it cannot guarantee it in general.

When explicit knowledge of the symmetries is available, incorporating equivariance into the model architecture provides stronger guarantees. Equivariant VAEs~\cite{Falorsi2018Reparameterizing}, for instance, explicitly enforce that latent variables transform according to the action of a symmetry group, ensuring alignment with symmetry directions by construction. Similarly, Lie Group VAEs~\cite{Falorsi2019Explorations} model latent spaces as manifolds structured by continuous group actions, allowing for more faithful symmetry representations.

A more intuitive argument stems from information compression. Symmetries in the dataset imply redundancies: data points related by symmetry transformations are not independent. Therefore, a model trained to reconstruct the data efficiently — as a VAE is — has an incentive to compress the representation by aligning latent directions with invariant or equivariant factors. This aligns with the observation that deep networks often favor simple internal representations when possible, a phenomenon sometimes referred to as the information bottleneck principle~\cite{Tishby2015DeepLearning}.

\subsection{A Toy Model: Linear VAE on Circular Data}

To illustrate these ideas in a tractable setting, consider a simple toy model. Let the data be sampled uniformly from a circle of fixed radius $R$ in two dimensions,
\begin{equation}
\mathcal{D} = \{ (x_1, x_2) \in \mathbb{R}^2 \, | \, x_1^2 + x_2^2 = R^2 \} \ ,
\end{equation}
as discussed in Sec.~\ref{sec:toy}.

Suppose we train a linear VAE — where both the encoder and decoder are linear maps — on this dataset. The optimal linear VAE will seek to represent the data efficiently under the ELBO objective. Because the data lies on a one-dimensional manifold (the circle), the model can minimize the reconstruction loss by using only one latent variable to parameterize the angular position along the circle, while the second latent dimension remains unused. Consequently, the latent space will self-organize so that one latent direction captures the essential degree of freedom (the angle), effectively aligning with the underlying $O(2)$ symmetry.

Let us show this behaviour explicitly. Assume a linear Variational Autoencoder (VAE), where the encoder and decoder are linear maps:
\begin{equation}
\mathbf{z} = W_{\text{enc}} \mathbf{x} \ , \quad \hat{\mathbf{x}} = W_{\text{dec}} \mathbf{z} \ .
\end{equation}

The VAE is trained to maximize the Evidence Lower Bound (ELBO), which, for a linear and deterministic encoder (neglecting stochastic noise for simplicity), reduces to minimizing the sum of the reconstruction loss and a KL regularization term encouraging $\mathbf{z}$ to match a standard Gaussian prior~\cite{Kingma2014VAE}.

The reconstruction loss is given by
\begin{equation}
\mathcal{L}_{\text{recon}} = \mathbb{E}_{\mathbf{x} \sim \mathcal{D}} \left[ \| \mathbf{x} - W_{\text{dec}} W_{\text{enc}} \mathbf{x} \|^2 \right] \ .
\end{equation}
Thus, the product $W_{\text{dec}} W_{\text{enc}}$ acts as an effective linear projection of $\mathbf{x}$ onto a low-dimensional subspace.

The dataset lies on a one-dimensional manifold (the circle), parametrized by an angle $\theta \in [0, 2\pi)$:
\begin{equation}
x_1 = R \cos \theta \ , \quad x_2 = R \sin \theta \ .
\end{equation}
Thus, the data is intrinsically one-dimensional despite being embedded in $\mathbb{R}^2$.

It is well known that the optimal linear approximation to a manifold, in the sense of minimizing reconstruction error, is obtained by projecting onto the principal subspace spanned by the leading principal components~\cite{Jolliffe2002PCA,Baldi1989LinearAE}. In this case, since the data is uniformly distributed over the circle, the covariance matrix of $\mathbf{x}$ is
\begin{equation}
\Sigma = \mathbb{E}_{\theta} \begin{pmatrix} x_1^2 & x_1 x_2 \\ x_1 x_2 & x_2^2 \end{pmatrix}
= \frac{R^2}{2} \begin{pmatrix} 1 & 0 \\ 0 & 1 \end{pmatrix} \ .
\end{equation}
Thus, $\Sigma$ is proportional to the identity matrix, reflecting the rotational symmetry of the circle.

Since $\Sigma$ is isotropic, any orthonormal basis is equally valid as a principal component basis. However, projecting onto any single direction in $\mathbb{R}^2$ captures only one-dimensional information, and the total variance captured is maximized by choosing any direction.

Therefore, the optimal linear encoder will map $\mathbf{x}$ onto a one-dimensional subspace, corresponding to a linear combination of $x_1$ and $x_2$, say
\begin{equation}
z = w_1 x_1 + w_2 x_2 \ ,
\end{equation}
with $w_1^2 + w_2^2 = 1$.

The KL regularization term in the VAE objective encourages the latent variable $z$ to have unit variance and zero mean, matching the standard Gaussian prior~\cite{Kingma2014VAE}. For data uniformly distributed over the circle, any linear projection onto a direction yields a distribution of $z$ that is approximately sinusoidal and thus approximately Gaussian for small segments.

Thus, the VAE is incentivized to use a single latent variable $z$ to encode the angular position $\theta$, while the second latent variable remains unused, corresponding to minimal KL divergence.

Minimizing the VAE objective leads to a solution where:
\begin{itemize}
    \item One latent dimension encodes the position along the circle (the angle $\theta$).
    \item The second latent dimension carries negligible information and is regularized to match the standard Gaussian prior.
\end{itemize}
Thus, the latent space self-organizes to align along the true intrinsic degree of freedom of the data — the symmetry direction corresponding to rotations along the circle.

Although this toy model relies on strong simplifications — linearity, perfect learning, noise-free data — it captures the essential intuition: the model benefits from compressing data along intrinsic symmetry directions to achieve efficient reconstruction.

In conclusion, while empirical evidence shows that VAEs often organize their latent spaces along symmetry directions when trained on symmetric data, a general proof for arbitrary datasets and models is not known. Partial theoretical results exist for linear models, disentangled VAEs, and equivariant architectures. In practice, the combination of inductive biases, strong regularization, and symmetry-induced redundancy in the data leads VAEs to favor latent representations that align with underlying symmetries.


\section{Conclusions}\label{sec:conclusions}

In this work, we explored how symmetries present in datasets manifest in the latent space structure learned by Variational Autoencoders (VAEs). By introducing a relevance measure for latent directions, we analyzed how the VAE self-organizes its latent space when trained on data with exact or approximate symmetries.

Using both a toy model and realistic physical processes, we showed that the presence of symmetries effectively reduces the intrinsic dimensionality of the data, and that this reduction is reflected in the organization of the latent space. In particular, datasets constrained by symmetry relations lead to a hierarchy in latent relevance, with a small number of latent variables capturing most of the meaningful variation. Our analysis of simple mechanical systems and particle collisions, including electron-positron and proton-proton scattering, confirmed that the latent structure aligns with the known physical constraints of the systems.

In addition to the empirical analysis, we provided a theoretical study of a simple toy model, demonstrating how, under idealized conditions, the latent space naturally aligns with the symmetry directions of the data manifold. This analysis supports the intuition that variational autoencoders, trained to balance reconstruction fidelity and compression, are incentivized to exploit symmetries present in the data.

These results highlight the potential of unsupervised generative models as tools for symmetry discovery in high-dimensional datasets. While our method does not require prior knowledge of the underlying symmetries, it is somewhat sensitive to model architecture and dataset characteristics, suggesting avenues for further refinement. In particular, extending this approach to more complex datasets, exploring different generative model architectures, and incorporating explicit regularization for symmetry properties are promising directions for future research.

Overall, our findings indicate that deep generative models not only learn to represent the data distribution but can also internalize fundamental structural properties, offering new possibilities for data-driven discovery in physics and beyond. Moreover, combining generative models with symmetry-aware architectures may enhance both discovery potential and interpretability, which could lead to new tools in data-driven science.

\section*{Aknowledgments}
 
The research of VS is supported by the Generalitat
Valenciana PROMETEO/2021/083, the Proyecto Consolidacion CNS2022-135688 from the AEI and the Ministerio de Ciencia e
Innovacion project PID2023-148162NB-C21, and the {\it Severo Ochoa} project CEX2023-001292-S funded by MCIU/AEI.

\bibliographystyle{JHEP}
\bibliography{bibtexsymms}

\end{document}